\documentclass[a4paper,twoside]{article}

\usepackage{epsfig}
\usepackage{subcaption}
\usepackage{calc}
\usepackage{amssymb}
\usepackage{amstext}
\usepackage{amsmath}
\usepackage{amsthm}
\usepackage{multicol}
\usepackage{pslatex}
\usepackage{apalike}
\usepackage{multirow}
\usepackage{algorithm2e}
\usepackage[bottom]{footmisc}
\usepackage{SCITEPRESS}     % Please add other packages that you may need BEFORE the SCITEPRESS.sty package.

\begin{document}

\title{Attention-based Shape and Gait Representations Learning for Video-based Cloth-Changing Person Re-Identification}

\author{\authorname{Vuong D. Nguyen, Samiha Mirza, Pranav Mantini, Shishir K. Shah}
% \author{\authorname{Vuong D. Nguyen\orcidAuthor{0000-0002-2369-8793}, Samiha Mirza\orcidAuthor{0000-0003-3754-6894}, Pranav Mantini\orcidAuthor{0000-0001-8871-9068} and Shishir K. Shah\orcidAuthor{0000-0003-4093-6906}}
\affiliation{Quantitative Imaging Lab, Dept. of Computer Science, University of Houston, Houston, Texas, USA}
\email{dnguy222@cougarnet.uh.edu, smirza6@uh.edu, pmantini@cs.uh.edu, sshah@central.uh.edu}
}

\abstract{Current state-of-the-art Video-based Person Re-Identification (Re-ID) primarily relies on appearance features extracted by deep learning models. These methods are not applicable for long-term analysis in real-world scenarios where persons have changed clothes, making appearance information unreliable. In this work, we deal with the practical problem of Video-based Cloth-Changing Person Re-ID (VCCRe-ID) by proposing ``\textbf{A}ttention-based
\textbf{S}hape and \textbf{G}ait Representations \textbf{L}earning'' (ASGL) for VCCRe-ID. Our ASGL framework improves Re-ID performance under clothing variations by learning clothing-invariant gait cues using a Spatial-Temporal Graph Attention Network (ST-GAT). Given the 3D-skeleton-based spatial-temporal graph, our proposed ST-GAT comprises multi-head attention modules, which are able to enhance the robustness of gait embeddings under viewpoint changes and occlusions. The ST-GAT amplifies the important motion ranges and reduces the influence of noisy poses. Then, the multi-head learning module effectively reserves beneficial local temporal dynamics of movement. We also boost discriminative power of person representations by learning body shape cues using a GAT. Experiments on two large-scale VCCRe-ID datasets
demonstrate that our proposed framework outperforms state-of-the-art
methods by $12.2\%$ in rank-1 accuracy and $7.0\%$ in mAP.}

\onecolumn \maketitle \normalsize \setcounter{footnote}{0} \vfill

\section{\uppercase{Introduction}}
\label{sec:introduction}
Person Re-Identification (Re-ID) involves matching the same person
across multiple non-overlapping cameras with variations in pose, lighting,
or appearance. Video-based Person Re-ID has been actively researched for various applications such as surveillance, unmanned tracking, search and rescue, etc. Two
main approaches have emerged: (1) Deep learning-based methods \cite{li2018cnnvideo,liu2019attvideo,gu2020ap3d}
that extract appearance features for Re-ID using Convolutional Neural Networks
(CNNs); and (2) Graph-based methods \cite{yang2020stvideo,wu2020agrl,khaldi2022} that
capture spatial-temporal information using Graph Convolutional Networks
(GCNs) \cite{kipf2017semisupervised}. However, these methods primarily rely on appearance features, making them likely to suffer performance degradation
in cloth-changing scenarios where texture information is unreliable.
This leads to a more practical Re-ID task called Cloth-Changing Person Re-ID
(CCRe-ID). 

Several methods have been proposed for image-based CCRe-ID, which
attempt to extract clothing-invariant modalities such as body shape
\cite{qian2020longterm,li2020shape}, contour sketches \cite{Yang_2021_prcc,Chen2022shape},
or silhouettes \cite{Hong2021shape}. Although these cues are more
stable than appearance in the long-term, extracting them from single-shot human image remains
challenging.
On the other hand, video-based data provides motion information that can improve matching ability of the Re-ID
system. Video-based CCRe-ID
(VCCRe-ID) has not been widely studied for two main reasons. First, there are only two public datasets: VCCR \cite{Han20223dshapevideo}, and CCVID \cite{Gu2022}
that are constructed from gait recognition datasets. Second,
capturing identity-aware cloth-invariant cues from video sequences
remains challenging in real-world scenarios. Texture-based works \cite{Gu2022,Cui2023Dcr-reid}
have proposed to extract clothing-unrelated features like faces or hairstyles.
However, these
methods fail under occlusion. Gait recognition models have been utilized \cite{Zhang20183dmotionvideo,Zhang2021cvid-reid} to assist the Re-ID systems.
However, these works do not efficiently capture the local temporal features
from video sequences. Moreover, they primarily rely on gait cues and
overlook identity-relevant shape features. These shortcomings necessitate
a more robust approach for VCCRe-ID. 

In this work, we propose ``\textbf{A}ttention-based \textbf{S}hape
and \textbf{G}ait Represenatations \textbf{L}earning'' (ASGL) framework for VCCRe-ID.
Our framework aims to mitigate the influence of clothing changes by
extracting texture-invariant body shape and gait cues
simultaneously from 3D skeleton-based human poses. The key components of AGSL are
the shape learning sub-branch and gait learning sub-branch, both of
which are built on Graph Attention Networks (GAT). The shape learning sub-branch is a GAT that processes a 3D skeleton sequences to obtain shape embedding, which is unique to individuals under clothing variations. The gait learning sub-branch is a Spatial-Temporal GAT (ST-GAT) that encodes gait from the skeleton-based spatial-temporal graph by modeling the temporal dynamics from movement of the body parts. This is different from previous works which
 leverage simple GCNs \cite{teepe2021gaitgraph,Zhang2021cvid-reid,khaldi2022}. The multi-head attention
mechanism in the proposed spatial-temporal graph attention blocks enables the framework to dynamically capture critical short-term movements by attending to important motion ranges in the sequence. This helps mitigate the influence of noisy frames caused by viewpoint changes or occlusion. We also reduce local feature redundancy in capturing motion patterns from pose sequence by narrowing the scope of self-attention operators, producing a discriminative gait embedding with beneficial information for Re-ID. Shape
and gait are then coupled with appearance for the global person
representation. 

In summary, our contributions in this work are as follows:
(1) we propose ASGL, a novel framework for the long-term
VCCRe-ID task; (2) we propose a ST-GAT for gait learning and a GAT
for shape learning, which helps to enhance the discriminative power
of identity representations under clothing variations and viewpoint
changes; and (3) we present extensive experiments on two large-scale
public VCCRe-ID datasets and demonstrate that our framework significantly
outperform state-of-the-art methods. 

\section{\uppercase{Related Works}}

\subsection{Person Re-ID}

Typically, early methods for image-based Re-ID include three main
approaches: representation learning \cite{Matsukawa2016ftlearning,Wang_2018},
metric learning \cite{Ma2014metric,Liao2015}, and deep learning \cite{Sun_2018_PCB,luo2019,Khaldi_2024_WACV}.
Video-based Re-ID methods focus on aggregating frame-wise appearance
features using 3D-CNN \cite{li2018cnnvideo,gu2020ap3d} and RNN-LSTM
\cite{Yichao2016rnnvideo,Zhen2017rnnvideo}, or capturing spatial-temporal
information using GNNs \cite{yang2020stvideo,wu2020agrl,khaldi2022}.
These methods produce comparable results on traditional person Re-ID
datasets \cite{Li2014cuhk03,Zheng2015market,zheng2016mars,li2018illisvid}.
However, these datasets were collected in short-term scenarios and the identities
present a consistency in appearance and clothing.  Existing methods trained on these datasets focus on encoding appearance features and hence are inefficient in long-term scenarios.

\subsection{Image-based CCRe-ID}

Recently, several datasets for image-based CCRe-ID have been published \cite{qian2020longterm,Yang_2021_prcc,wan2020vc}. Huang \textit{et al.} \cite{Huang2020} proposed
to capture clothing variations within the same identity using vector-neuron
capsules. Body shape cues are explicitly extracted by Qian \textit{et al.}
\cite{qian2020longterm} using a shape-distillation module, and by
Li \textit{et al.} \cite{li2020shape} using adversarial learning. Other works also attempt to extract modalities that
are stable in long-term, such as 2D skeleton-based poses \cite{Nguyen_2024_CVSL} silhouettes \cite{Hong2021shape,jin2022gait}, or contour sketches \cite{Yang_2021_prcc}.
These works mostly rely on , which are affected
by viewpoint changes, making extracted features ambiguous for Re-ID.
Chen \textit{et al.} \cite{Chen20213dshapeimage} proposed to estimate and regularize
3D shape parameters using projected 2D pose and silhouettes. Zheng
\textit{et al.} \cite{Zheng_20223dimage} leveraged 3D mesh to jointly learnt
appearance and shape features. However, image-based setting is sensitive
to the quality of Re-ID data and less tolerant to noise due to limited
information contained in a single person image.

\begin{figure*}[t]
    \centering
    \includegraphics[width=\textwidth]{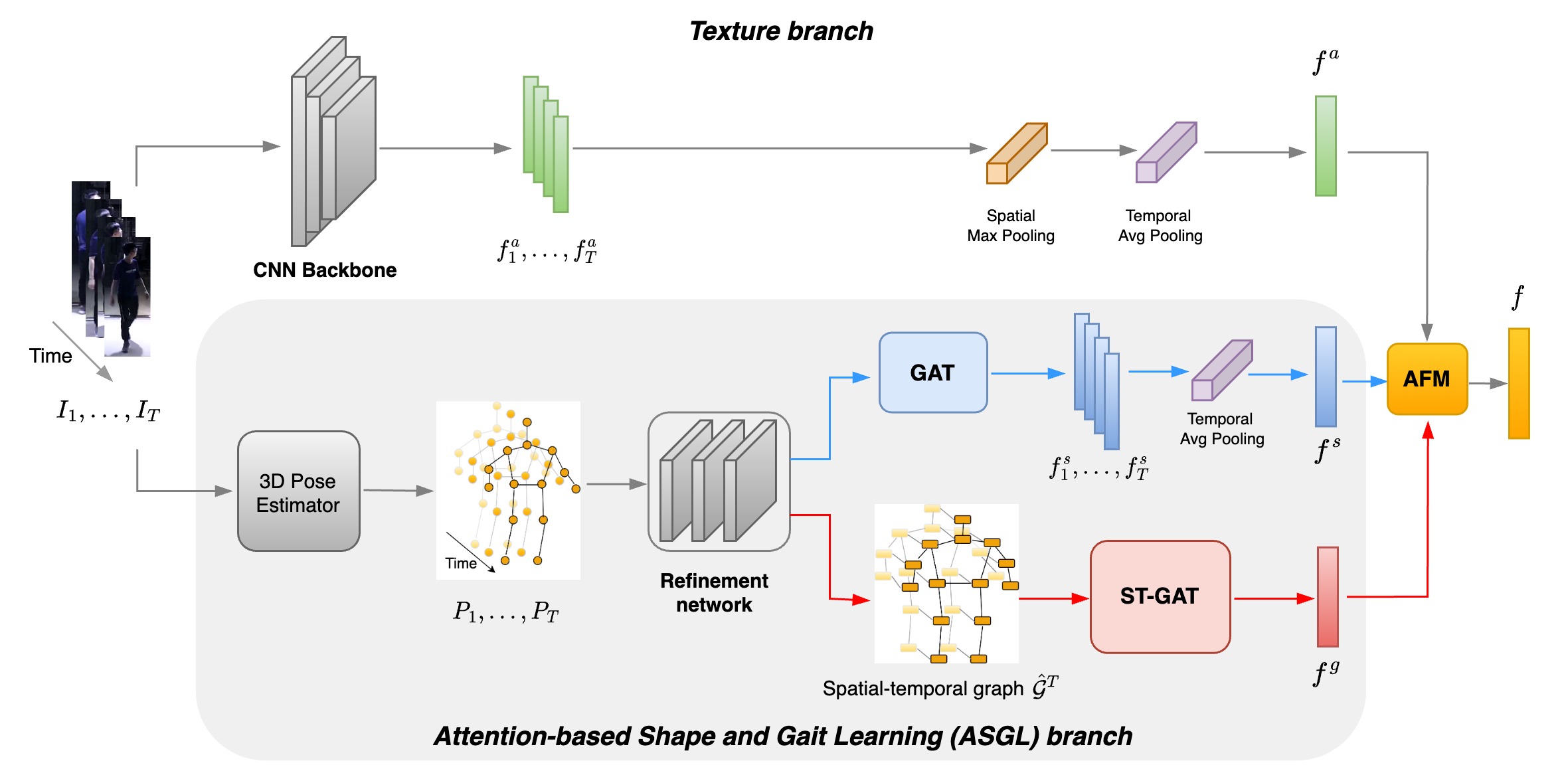}
    \caption{Overview of the proposed ASGL framework. Given a video sequence, for the ASGL branch, 3D pose sequence is first estimated and then refined. A GAT in shape learning sub-branch extracts frame-wise shape features, which are then aggregated for the video-wise shape embedding by a temporal average pooling layer (blue flow). Meanwhile, a spatial-temporal graph is constructed from the refined pose sequence, which is then processed by a ST-GAT to obtain gait embedding (red flow). Appearance, shape and gait are finally fused by the Adaptive Fusion module for the final person representation.}
    \label{fig:baseline}
\end{figure*}

\subsection{Video-based CCRe-ID}

Video-based CCRe-ID has not been widely studied due to the limited
number of publicly available datasets. Previous works on VCCRe-ID
can be categorized into two main approaches. First, texture-based
methods, where Gu \textit{et al.} \cite{Gu2022} designed clothes-based losses to
eliminate the influence of clothes on global appearance. Bansal \textit{et al.} \cite{Bansal2022self-att} and Cui \textit{et al.} \cite{Cui2023Dcr-reid}
leveraged self-attention to attend to appearance-based cloth-invariant
features like face or hairstyle. These methods suffer severe performance
degradation under occlusion.
Second, gait-based methods \cite{Zhang20183dmotionvideo,Zhang2021cvid-reid,wang2022sccvreid}, where
motion patterns are captured as features for Re-ID. These works first
assume constant walking trajectory from identities, which is not practical
in real-world. Then gait cues are encoded from sequences of 2D poses
by GNNs. There are two limitations to this approach:
(1) viewpoint changes significantly limit the ability to capture body parts
movement from 2D pose; and (2) simple GNNs do not capture local
motion patterns efficiently. Recent works \cite{Han20223dshapevideo,Nguyen_2024_SEMI} capture 3D SMPL shape to mitigate viewpoint changes. For example, Han \textit{et al.} \cite{Han20223dshapevideo} proposed to
extract human 3D shape cues using a two-stage framework, which needs
additional large-scale datasets and results in a heavy training process. In this work, we address these issues by simultaneously
extracting body shape along with gait from 3D pose using attention-based
variations of GNNs.

\section{\uppercase{Proposed framework}}

\subsection{Overview}

An overview of the proposed ASGL framework is demonstrated in Figure
\ref{fig:baseline} Given a $T$-frame video sequence $\{I_{i}\}_{i=1}^{T}$ as input,
we first employ an off-the-shelf 3D pose estimation model to estimate
frame-wise 3D pose sequence $P=\{P_{i}\}_{i=1}^{T}$. $P$ is then
fed into the refinement network $\mathcal{R}(.)$, giving refined
sequence of frame-wise skeleton-based features $\hat{J}=\left\{ \hat{J}_{P_{i}}\right\} _{i=1}^{T}$.
The shape learning sub-branch which comprises of a Graph Attention
Network extracts frame-wise shape feature vectors $\{f_{i}^{s}\}_{i=1}^{T}$
from $\hat{J}$, then aggregates $\{f_{i}^{s}\}_{i=1}^{T}$ for the
video-wise shape embedding $f^{s}$ using a temporal average pooling
layer. Meanwhile, the gait learning sub-branch first connects $\hat{J}$
to yield the spatial-temporal motion graph $\mathcal{G}^{st}$, then
uses the proposed Spatial-Temporal Graph Attention Network to encode
gait embedding $f^{g}$ from $\mathcal{G}^{st}$. Since texture information
is still important in the cases of slight clothing changes, we extract
appearance embedding $f^{a}$ using a CNN backbone. Finally, $f^{a},f^{s}$
and $f^{g}$ are fused by the Adaptive Fusion Module for the final
person representation $f$, which is then fed into the cross-entropy loss and pair-wise triplet loss functions for training the framework. In testing stage, matching is performed by comparing the video-wise representations based on cosine distance.

\subsection{Attention-based Shape and Gait Learning branch}

The goal of the Attention-based Shape and Gait Learning (ASGL) branch
is to learn body shape and gait features, which serve as complementary information to appearance features for a robust person representation in long-term scenarios. ASGL comprises of a 3D pose estimator, a refinement
network to refine the frame-wise skeleton-based pose sequence, a shape learning sub-branch built on a GAT and a gait learning sub-branch built on a ST-GAT.

\subsubsection{Pose Estimator and Refinement Network}

\begin{figure} [t]
    \centering
    \includegraphics[width=0.75\columnwidth]{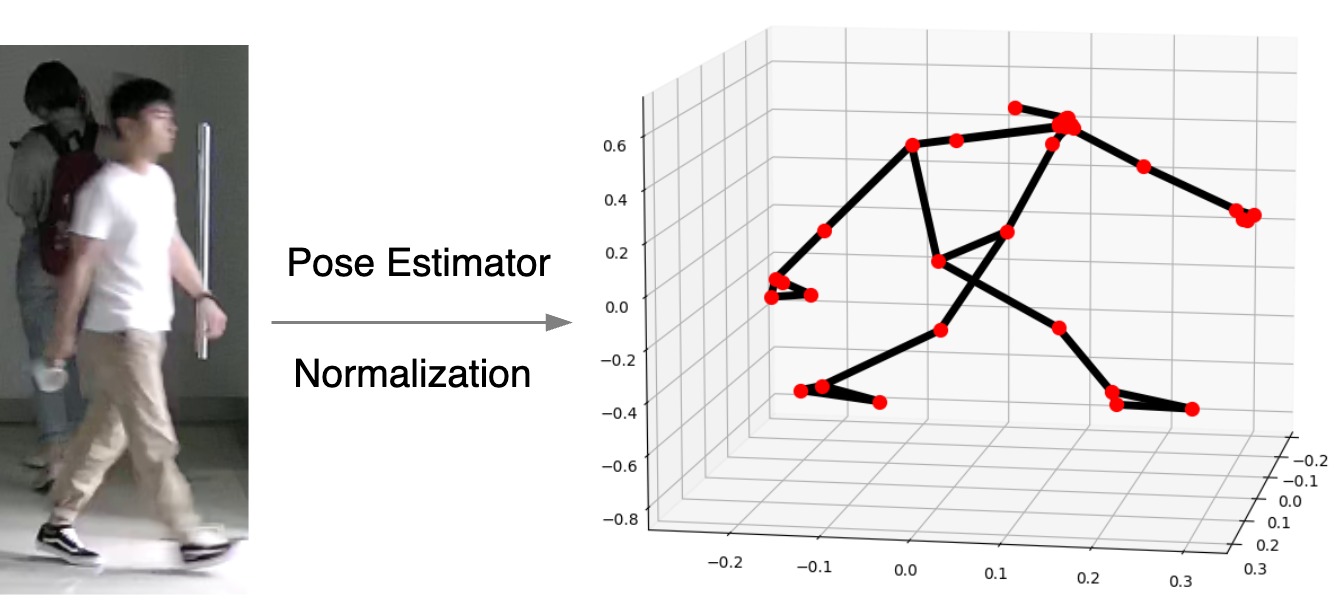}
    \caption{Illustration of 3D pose estimation. Pose is first estimated using an off-the-shelf pose estimator, then normalized to an unified view.}
    \label{fig:sample_pose}
\end{figure}

In contrast to existing methods \cite{qian2020longterm,teepe2021gaitgraph}
that use 2D pose, we utilize 3D pose for learning shape
and gait. 3D pose is more robust to
camera viewpoint changes and occlusions. We employ an off-the-shelf 3D pose
estimator \cite{bazarevsky2020blazepose} to obtain frame-wise pose sequence $P=\{P_{i}\}_{i=1}^{T}$.
The joint set $J_{P_{i}}=\{j_{i}\}_{i=1}^{k}$ corresponding to pose
$P_{i}$ contains $k$ 3D keypoints as illustrated in Figure \ref{fig:sample_pose}.
Each estimated keypoint $j_{i}$ is represented as a set of three
coordinates $(x_{i},y_{i},z_{i})$ indicating the location of certain
body parts. 

To avoid misalignment in capturing motion patterns caused by camera
viewpoint variations, for every keypoint $j_{i}\in J_{P_{i}}$, we
first translate the raw keypoints to the origin of coordinate: 
\begin{equation}
(x_{i},y_{i},z_{i})\rightarrow(x_{i}+\Delta_{i,x},y_{i}+\Delta_{i,y},z_{i}+\Delta_{i,z})
\end{equation}
where $(\Delta_{i,x},\Delta_{i,y},\Delta_{i,z})$ is the translation
offset. We then normalize the translated keypoints to an unified view
as follows: $\overline{j}_{i}=\left(\frac{x_{i}+\Delta_{i,x}}{h},\frac{y_{i}+\Delta_{i,y}}{w},\frac{z_{i}+\Delta_{i,z}}{h\times w}\right)$,
where $(h,w)$ is the size of input frame. The normalized keypoint
set is then refined to obtain $\hat{j}_{i}\in\mathbb{R}^{d}$ using
the refinement network $\mathcal{R}(\cdot)$, i.e. $\hat{j}_{i}=\mathcal{R}\left(\overline{j}_{i}\right)$.
$\mathcal{R}(\cdot)$ consists of a sequence of fully connected layers,
which aims to capture fine-grained details of the person's body shape
via the high-dimensional keypoint-wise feature vector set $\hat{J}_{P_{i}}=\left\{ \hat{j}_{i}\right\} _{i=1}^{k}$.
Output of the refinement network is the refined frame-wise pose sequence
$\hat{J}=\left\{ \hat{J}_{P_{i}}\right\} _{i=1}^{T}$.

\subsubsection{Shape Representation Learning}

Body shape remains relatively stable in long-term, thus it can serve
as an important cue for CCRe-ID. Using the 3D skeleton representation,
shape describes the geometric form of the human body. Intuitively,
body shape can not be captured via a single keypoint-wise feature
vector. In this work, we propose a Graph Attention Network (GAT) \cite{velickovic2017graph}
to model the relations between pairs of connected keypoints. GAT is
a type of GCN, which uses message passing to learn features across
neighborhoods from the skeleton-based graph. GAT applies an attention
mechanism in the aggregation and updating process across several graph
attention layers. This helps to exploit the local relationships between
body parts for a discriminative shape embedding of the person. 

Specifically, a graph that represents the body pose is first constructed
from the refined keypoint set $\hat{J}_{P_{i}}=\left\{ \hat{j}_{i}\right\} _{i=1}^{k}$,
in which keypoints are nodes and bones are edges of the graph. Each
node $\hat{j}_{i}$ corresponds to a set $Q_{i}$ containing indices
of neighbors of $\hat{j}_{i}$. $Q_{i}$ can be constructed via adjacency
matrix. Then, the $l^{th}$ layer of GAT $\mathcal{G}$ updates $\hat{j}_{i}$
by aggregating information from $Q_i$, given as:

\begin{equation}
\hat{j}_{i}^{(l+1)}=\sigma\left(\sum_{j\in Q_{i}}\mathbf{W}_{ij}\theta^{(l)}\hat{j}_{i}^{(l)}\right)
\end{equation}
where $\mathbf{W}=(a_{ij}),\mathbf{W}\in\mathbb{R}^{k\times k}$,
$a_{ij}$ stores the weighting between $\hat{j}_{i}$ and $\hat{j}_{j}$
(i.e. the importance of joint $\hat{j}_{j}$ to joint $\hat{j}_{i}$).
$\theta^{(l)}\in\mathbb{R}^{d^{(l+1)}\times d^{(l)}}$ is the weight
matrix of the $l^{th}$ graph attention layer $\mathcal{G}^{(l)}$,
where $d^{(l)}$ is the dimension of layer $\mathcal{G}^{(l)}$. $\sigma$
is an activation function. GAT implicitly amplifies importances of
each joint to its neighbors. To do this, unlike traditional
GCNs in which $\mathbf{W}$ is explicitly defined, GAT $\mathcal{G}$
implicitly computes $a_{ij}\in\mathbf{W}$ by: 
\begin{equation}
a_{ij}=\text{softmax}_{j}\,\,h\left(\mathbf{\theta}\hat{j}_{i},\mathbf{\theta}\hat{j}_{j}\right)
\end{equation}
 where $h:\mathbb{R}^{d^{(l+1)}}\times\mathbb{R}^{d^{(l+1)}}\rightarrow\mathbb{R}$
is a byproduct of an attentional mechanism. We employ a global max
pooling layer to aggregate the higher-order representations of joint
set $\hat{J}_{P_{i}}^{(L-1)}$ after $L$ GAT layers for the frame-wise
shape embedding $f_{i}^{s}$ of the $i^{th}$ frame as follows:
\begin{equation}
f_{i}^{s}=\text{GMP}\left(\hat{J}_{P_{i}}^{(L-1)}\right)
\end{equation}
where $\text{GMP}$ denotes global max pooling and summarizes the
discriminative information in the graph. The frame-wise shape embedding
set $\{f_{i}^{s}\}_{i=1}^{T}$ is finally fed into a temporal average
pooling layer to obtain the video-wise shape representation $f^{s}$.

\subsubsection{Gait Representation Learning }

\begin{figure*} [t]
    \centering
    \includegraphics[width=0.75\textwidth]{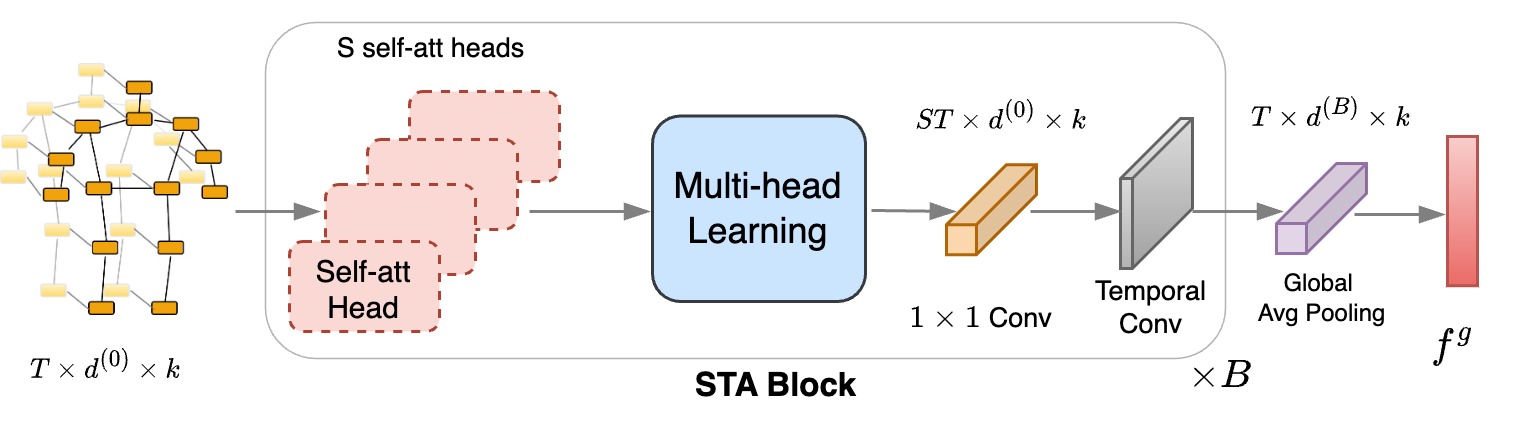}
    \caption{Architecture of the proposed Spatial-Temporal Graph Attention Network for encoding skeleton-based gait.}
    \label{fig:stgat}
\end{figure*}

Unlike previous works \cite{teepe2021gaitgraph,Zhang2021cvid-reid}
that encode gait using traditional variations of Spatial-Temporal
Graph Convolutional Networks (ST-GCN) \cite{yan2018stgcn}, we propose to
learn gait cues using a Spatial-Temporal Graph Attention Network (ST-GAT)
$\mathbf{G}_{sta}$. By incorporating an attention mechanism, we enable
the network to effectively amplify the important motion patterns and reduce the influence of noisy motion ranges, producing gait representations with high discriminative power.

As shown in Figure \ref{fig:stgat}, input for the gait learning sub-branch is
the spatial-temporal graph $\hat{\mathcal{G}}^{T}\in\mathbb{R}^{T\times k\times d}$
constructed from the refined skeleton sequence $\hat{J}=\left\{ \hat{J}_{P_{i}}\right\} _{i=1}^{T}$.
In this work, we follow the spatial-temporal connection as in \cite{yan2018stgcn},
allowing for direct aggregation of moving information from consecutive
frames. Our ST-GAT $\mathbf{G}_{sta}$ consists of $B$ consecutive
Spatial-Temporal Attention (STA) blocks, followed by a global average
pooling to output gait representation $f^{g}$. For each STA block,
we first employ multi-head attention modules $\mathcal{H}$. As occlusions
may lead to noisy frames with unobservable movements of body parts,
$\mathcal{H}$ allows to attend to different ranges of motion patterns
simultaneously. This helps amplify the most contributing frame-wise
skeletons to the global gait encoding and reduce the influence of
noisy frames. In this work, $\mathcal{H}$ consists of $S$ independent
self-attention modules, which capture the spatial-temporal dynamics
of the input spatial-temporal graph:
\begin{equation}
\hat{\mathcal{G}}^{att}=\sigma\left(\frac{1}{S}\sum_{s=1}^{S}W_{s}\hat{\mathcal{G}}^{T}\mathbf{A}_{s}\right)
\end{equation}
where $W_{s}$ is the learnable attention matrix of the $s^{th}$
head which weights the edge importance, i.e. the relationships among
connected joints, $\mathbf{A}_{s}$ is the adjacency matrix of the
$s^{th}$ head, $\sigma$ is the activation function and $\hat{\mathcal{G}}^{att}$
is the accumulated output of all heads. 

$\hat{\mathcal{G}}^{att}$ is then fed into the multi-head learning
module which consists of several $1\times1$ convolutional layers.
Compared to the single-frame spatial graph, the size of the spatial-temporal
graph increases $T$ times. This limits the ability of the self-attention
operators to adaptively construct relationships between joints and
neighbors. Therefore, in the multi-head learning module, we first
partition the graph into several small groups to limit the number
of neighbor nodes for each joint:

\begin{equation}
\mathcal{N}\left(\hat{j}_{i}\right)=\left\{ \hat{j}_{j}\,|\,d\left(\hat{j}_{i},\hat{j}_{i}\right)\le D\right\} 
\end{equation}
where $\mathcal{N}$ is the neighbor set, $d(\cdot,\cdot)$ denotes
the shortest graph distance between two nodes. In this work, we set
$D=3$. Then, the ST-GAT $\mathbf{G}_{sta}$ only weighs edges within
groups, thus helps to reserve beneficial local motion patterns and
reduce computation costs.

STA block then aggregates the temporal information learnt by $S$
self-attention heads using a temporal convolutional layer. The coarse-grained
gait encoding after the $B^{th}$ STA blocks is finally summarized by a global
average pooling layer, giving the fine-grained video-wise gait representation
$f^{g}$.

\subsection{Adaptive Fusion Module}

When the identities slightly change clothes, appearance remains competitive in visual similarities. Frame-wise appearance feature set is first extracted by a CNN backbone, then aggregated by a spatial max pooling and
temporal average pooling to obtain the video-wise appearance embedding
$f^{a}$ (Figure \ref{fig:baseline}). 

\begin{figure} [ht]
    \centering
    \includegraphics[width=\columnwidth]{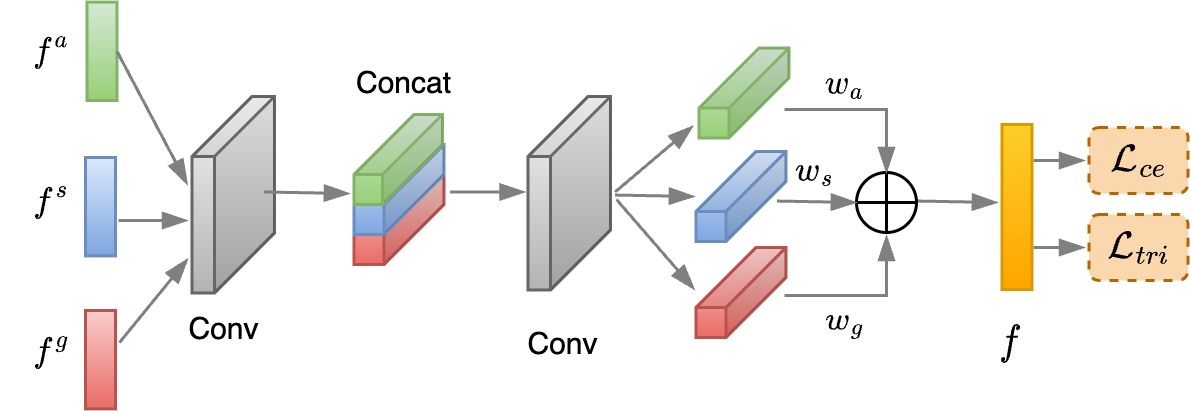}
    \caption{Architecture of the Adaptive Fusion Module.}
    \label{fig:afm}
\end{figure}

We finally fuse appearance, shape, and gait embeddings using the Adaptive Fusion Module (AFM) as illustrated in Figure \ref{fig:afm}. Embeddings are first projected onto a common latent space. Then, they are stacked using concatenation and fed into a convolutional layer which aims to optimize the embeddings in parallel by making them refer to each other. The corresponding weights $w_a$, $w_s$ and $w_g$ are estimated to adaptively amplify the importance of each embedding for the final person representation $f$. This is useful since viewpoint changes and occlusions bring different level of semantic information from appearance, gait, and shape for certain input videos.

Our ASGL framework is supervised by the sum of two identification loss functions:
\begin{equation}
\mathcal{L}=\lambda_{1}\mathcal{L}_{ce}+\lambda_{2}\mathcal{L}_{tri}
\end{equation}
where $\mathcal{L}_{ce}$ is cross-entropy loss, $\mathcal{L}_{tri}$
is pair-wise triplet loss and $\lambda_{1},\lambda_{2}$
are weighting parameters.

\section{\uppercase{Experimental Setup}}
\subsection{Datasets and Evaluation Protocols}

\begin{figure}[t]
\begin{centering}
\includegraphics[width=\columnwidth]{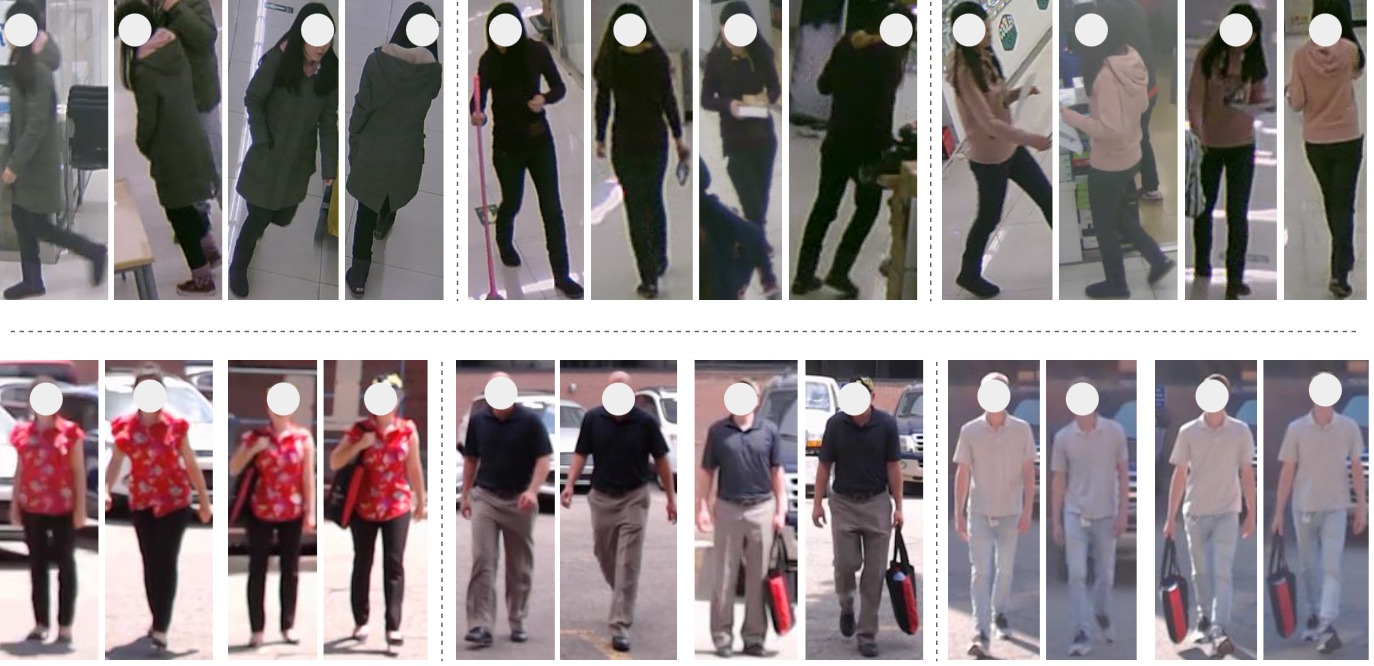}
\par\end{centering}
\caption{\label{datasets_comparison}Samples from VCCR (top) and CCVID (bottom). For VCCR, we randomly collect $3$ tracklets from the \textbf{same identity} under different clothing. For CCVID, we randomly choose $3$ identities with $2$ tracklets each under different clothing. VCCR poses realistic challenges for Re-ID like entire clothing changes, viewpoint variations, and occlusions, while CCVID contains only frontal images, clearly visible faces, no occlusion and slight clothing change with identities carrying bags.
}
\end{figure}

\textbf{Datasets:} We validate the performance of our proposed framework on two public VCCRe-ID datasets, VCCR \cite{Han20223dshapevideo}
and CCVID \cite{Gu2022}. \textbf{VCCR} contains
$4,384$ tracklets with $392$ identities, with $2\sim10$
different suits per identity. \textbf{CCVID} contains $2,856$ tracklets with $226$ identities, with $2\sim5$
different suits per identity. In Figure \ref{datasets_comparison}, we report a relative comparison in challenges for Re-ID posed by the two datasets by showing samples randomly selected and viewpoint variations. It can be seen that CCVID mimics simplistic Re-ID scenarios such as frontal viewpoints, clearly visible faces, or no occlusion, while VCCR poses complex real-world scenarios for Re-ID. Therefore, we focus on validating the effectiveness of our framework on VCCR. \par

\paragraph{Evaluation protocols:} Two Re-ID metrics are used to evaluate the effectiveness of our method: Rank-k (R-k) accuracy and mean Average Precision
(mAP). We compute
testing accuracy in three settings: (1) \textbf{Cloth-Changing} (CC), i.e.
the test sets contains only cloth-changing ground truth samples; (2)
\textbf{Standard}, i.e. the test sets contain both cloth-changing and cloth-consistent
ground truth samples; and (3) \textbf{Same-clothes} (SC), i.e. the test sets contain only
cloth-consistent ground truth samples. 

\subsection{Implementation Details}

\paragraph{Model Architecture.} We choose Resnet-50 \cite{he2015resnet50} pretrained
on ImageNet \cite{Deng2009imagenet} as the CNN backbone for texture
branch. For the ASGL branch, we employ MediaPipe \cite{bazarevsky2020blazepose}, an off-the-shelf estimator 
for 3D pose estimation, which outputs $33$ keypoints in world coordinates.
As we focus on capturing moving patterns of major joints and do not
need features from noses, eyes, fingers, or heels. We average keypoints
on face, hand and foot to single keypoints, leaving the skeleton graph
with $14$ keypoints. The refinement networks consists of $3$ fully
connected layers of size $[128,512,2048]$ respectively. For shape learning
sub-branch, the GAT consists of two graph attention layers. For gait
learning sub-branch, our ST-GAT consists of $2$ STA blocks with their
channels in $[128,256]$. Implementation is in PyTorch \cite{paszke2019pytorch}.

\paragraph{Training and Testing.} Input clips for training and testing are formed
by randomly sampling $8$ frames from each original tracklet with a
stride of $2$ for VCCR and $4$ for CCVID. Frames are resized to
$256\times128$, then horizontal flipping is applied for data augmentation. The batch size
is set to $32$, each batch randomly selects $8$ identities and $4$
clips per identity. The model is trained using Adam \cite{kingma2017adam}
optimizer for $120$ epochs. Learning rate is initialized at $5\times e^{-3}$
and reduced by a factor of $0.1$ after every $40$ epochs. We set $\lambda_1=0.7, \lambda_2=0.3$.

\section{\uppercase{Experimental Results}}

\begin{table*}
\small
\caption{\label{tab:vccr_results}Quantitative results on VCCR. ASGL outperforms SOTAs
in all evaluation settings by a significant margin}
\begin{centering}
\begin{tabular}{c|c|c|cc|cc|cc}
\hline 
\multirow{2}{*}{Method} & \multirow{2}{*}{Method type} & \multirow{2}{*}{Modalities} & \multicolumn{2}{c|}{CC} & \multicolumn{2}{c|}{Standard} & \multicolumn{2}{c}{SC}\tabularnewline
\cline{4-9} \cline{5-9} \cline{6-9} \cline{7-9} \cline{8-9} \cline{9-9} 
 &  &  & R-1 & mAP & R-1 & mAP & R-1 & mAP\tabularnewline
\hline 
\hline 
PCB \cite{Sun_2018_PCB} & Image-based & RGB & $18.8$ & $15.6$ & $55.6$ & $36.6$ & - & -\tabularnewline
AP3D \cite{gu2020ap3d} & Video-based & RGB & $35.9$ & $31.6$ & $78.0$ & $52.1$ & - & -\tabularnewline

GRL \cite{Liu2021} & Video-based & RGB & $35.7$ & $31.8$ & $76.9$ & $51.4$ & - & -\tabularnewline
\hline 
SPS \cite{Shu_2021_SPS} & Image-based CC & RGB & $34.5$ & $30.5$ & $76.5$ & $50.6$ & - & -\tabularnewline
CAL \cite{Gu2022} & Video-based CC & RGB & $36.6$ & $31.9$ & $78.9$ & $52.9$ & $79.1$ & $63.8$\tabularnewline

3STA \cite{Han20223dshapevideo} & Video-based CC & RGB + 3D shape & $40.7$ & $36.2$ & $80.5$ & $54.3$ & - & -\tabularnewline
\hline 
\textbf{ASGL (Ours)} & Video-based CC & RGB + shape + gait & $\mathbf{52.9}$ & $\mathbf{43.2}$ & $\mathbf{88.1}$ & $\mathbf{65.8}$ & $\mathbf{89.9}$ & $\mathbf{79.5}$\tabularnewline
\hline 
\end{tabular}
\par\end{centering}

\end{table*}

\begin{table*}[t]
\small
\caption{\label{ccvid_results}Comparison of quantitative results on CCVID.}
\begin{centering}
\begin{tabular}{c|cc|cc}
\hline 
\multirow{2}{*}{Method} & \multicolumn{2}{c|}{CC} & \multicolumn{2}{c}{Standard}\tabularnewline
\cline{2-5} \cline{3-5} \cline{4-5} \cline{5-5} 
 & R-1 & mAP & R-1 & mAP\tabularnewline
\hline 
\hline 
InsightFace \cite{Deng2020Retina} & $73.5$ & - & $\mathbf{95.3}$ & -\tabularnewline

ReFace \cite{arkushin2022reface} & $\mathbf{90.5}$ & - & $89.2$ & -\tabularnewline
\hline
CAL \cite{Gu2022} & $81.7$ & $79.6$ & $82.6$ & $81.3$\tabularnewline

DCR-ReID \cite{Cui2023Dcr-reid} & $83.6$ & $81.4$ & $84.7$ & $82.7$\tabularnewline
\hline
\textbf{ASGL (Ours)} & $83.9$ & $\mathbf{82.2}$ & $86.1$ & $82.5$\tabularnewline
\hline 
\end{tabular}
\par\end{centering}
\end{table*}

\subsection{Results on VCCR}

A comparison of the quantitative results on VCCR \cite{Han20223dshapevideo}
dataset is reported in in Table \ref{tab:vccr_results}. State-of-the-art results categorized
by method types are presented as benchmark.  These include image-based short-term
Re-ID (i.e. PCB \cite{Sun_2018_PCB}), video-based short-term Re-ID
(i.e. AP3D \cite{gu2020ap3d} and GRL \cite{Liu2021}), image-based
CCRe-ID (i.e. SPS \cite{Shu_2021_SPS}) and video-based CCRe-ID (i.e.
CAL \cite{Gu2022} and 3STA \cite{Han20223dshapevideo}). 

Overall, ASGL outperforms previous methods on VCCR in all evaluation
protocols. Compared to image-based methods, our method achieves better
performance, indicating the importance of spatial-temporal information
for Re-ID. The texture-based method CAL \cite{Gu2022} is outperformed
by our framework by $16.3\%$ in rank-1 accuracy and $11.3\%$ in mAP in
cloth-changing setting, which shows the effectiveness of ASGL in coupling
auxiliary modalities (i.e. shape and gait) with appearance for VCCRe-ID.
In same-clothes setting, which mimics the short-term Re-ID environment,
ASGL is also superior to CAL, which demonstrates the robustness of
ASGL in real-world scenarios. It can be reasoned that severe occlusion
and viewpoint changes posed by VCCR hinders the ability of CAL to
capture face and hairstyle.

Compared to the 3STA \cite{Han20223dshapevideo} framework, in cloth-changing
setting, ASGL achieves a remarkable improvement of $12.2\%$ in rank-1 and $7.0\%$ in mAP. It can be seen that the combination
of shape and gait cues significantly improve the discriminative power
of person representations. Moreover, 3STA framework is multi-stage
and requires heavy training, shown by a number of $250$ training
epochs for the first stage and $30000$ training epochs for the second
stage as reported in \cite{Han20223dshapevideo}. Our framework instead
can be trained in an end-to-end manner with only $120$ epochs. 

\subsection{Results on CCVID}

On CCVID \cite{Gu2022} dataset, we report experimental results of our method along with two texture-based methods
including CAL \cite{Gu2022} and DCR-ReID \cite{Cui2023Dcr-reid} in Table \ref{ccvid_results}. It can be seen that the models that focus on extracting
appearance features achieve comparable performance on CCVID. Moreover, the results are close to saturation. The limitation of these methods is that they rely heavily on the assumption that input frames contain clearly visible persons. Importantly, CCVID mimics unrealistic Re-ID environment, in which all identities walk towards camera, giving frontal viewpoint, no occlusion. Clothing variations only include carrying a bag or wearing a cap, leading to very slight clothing changes. Therefore, we do not focus on validating the effectiveness of our ASGL framework on CCVID. 

\subsection{Ablation study}

In ablation study, we validate the effectiveness of: (1) shape and
gait embeddings produced by the Attention-based Shape and Gait Learning (ASGL) branch; (2) the proposed GAT
in modeling shape and ST-GAT in modeling gait compared to traditional GCN and ST-GCN; and (3) using 3D pose
compared to 2D pose.

\begin{table*}[t]
\begin{centering}
\small
\caption{\label{tab:AS_shape}Ablation study of the ASGL branch on VCCR and
CCVID.}
\begin{tabular}{c|cc|cc|cc|cc}
\hline 
\multirow{3}{*}{Method} & \multicolumn{4}{c|}{VCCR} & \multicolumn{4}{c}{CCVID}\tabularnewline
\cline{2-9} \cline{3-9} \cline{4-9} \cline{5-9} \cline{6-9} \cline{7-9} \cline{8-9} \cline{9-9} 
 & \multicolumn{2}{c|}{CC} & \multicolumn{2}{c|}{Standard} & \multicolumn{2}{c|}{CC} & \multicolumn{2}{c}{Standard}\tabularnewline
\cline{2-9} \cline{3-9} \cline{4-9} \cline{5-9} \cline{6-9} \cline{7-9} \cline{8-9} \cline{9-9} 
 & R-1 & mAP & R-1 & mAP & R-1 & mAP & R-1 & mAP\tabularnewline
\hline 
\hline 
Texture branch (Appearance) & $32.8$ & $29.3$ & $74.3$ & $46.7$ & $78.5$ & $75.3$ & $79.7$ & $76.9$\tabularnewline
ASGL branch (Shape and Gait) & $29.1$ & $27.4$ & $68.2$ & $43.9$ & $71.3$ & $70.3$ & $72.1$ & $70.8$\tabularnewline
\hline 
Appearance and Shape & $38.7$ & $32.3$ & $77.2$ & $50.7$ & $79.6$ & $75.6$ & $79.9$ & $77.2$\tabularnewline
Appearance and Gait & $41.5$ & $35.5$ & $79.6$ & $53.1$ & $79.2$ & $76.1$ & $80.6$ & $77.9$\tabularnewline
\hline 
Joint (the proposed ASGL) & $\mathbf{52.9}$ & $\mathbf{43.2}$ & $\mathbf{88.1}$ & $\mathbf{65.8}$ & $\mathbf{83.9}$ & $\mathbf{82.2}$ & $\mathbf{86.1}$ & $\mathbf{82.5}$\tabularnewline
\hline 
\end{tabular}
\par\end{centering}

\end{table*}

\paragraph{ASGL Branch.} To validate the effectiveness of the various modules in the proposed framework, we carried out training with the following
model settings: Texture branch (only appearance embeddings are extracted),
ASGL branch (only shape and gait embeddings are extracted), and joint
representations. The experimental results are reported in Table \ref{tab:AS_shape}.

It can be observed that the model using only ASGL branch performs
worse than that using only Texture branch on VCCR with a gap of $3.7\%$ in rank-1 accuracy and $1.9\%$ in mAP in cloth-changing setting.
The reasons is that appearance features are still more competitive
in visual similarities than pose-based modalities when identities do not change or slightly change clothes. Moreover, the discriminative power of shape and gait embeddings rely on the accuracy and robustness of the off-the-shelf pose estimator, which may suffer poor estimation results under severe occlusions. Overall, the matching ability of our Re-ID framework is maximized when appearance is coupled with shape and gait
embeddings extracted by ASGL branch, shown by a large performance
gap of $20.1\%/23.8\%$ in rank-1 accuracy and $13.9\%/15.8\%$ in mAP between the
joint model and the single-branch texture/ASGL models. 

We further analyze the contribution of each pose-based cue (i.e. shape
and gait) to the discriminability of person representations, in which
two models are trained: appearance coupled with shape and appearance
coupled with gait. On VCCR, appearance-gait model achieves higher rank-1
accuracy in both evaluation protocols than appearance-shape model,
while on CCVID, there is no significant difference in performance
between two models. This can be reasoned by the challenges posed by
the two datasets. Video tracklets in CCVID only contain frontal walking
trajectories (i.e. people walking towards camera) and no occlusion, while VCCR poses greater viewpoint variations and occlusions. Thus,
VCCR brings richer gait information and its nature hinders the extraction
of fine-grained shape embeddings. It is also worth noting that coupling
both shape and gait cues with appearance brings the most discriminative
power for Re-ID in real-world scenarios, shown by the results of our
proposed ASGL framework.

\begin{table}[h]
\caption{\label{tab:GAT}Ablation study of the proposed GAT for learning shape and ST-GAT for learning gait on VCCR.}
\begin{centering}
\small
\begin{tabular}{c|cc|cc}
\hline 
\multirow{2}{*}{Method} & \multicolumn{2}{c|}{CC} & \multicolumn{2}{c}{Standard}\tabularnewline
\cline{2-5} \cline{3-5} \cline{4-5} \cline{5-5} 
 & R-1 & mAP & R-1 & mAP\tabularnewline
\hline 
\hline 
GCN \& ST-GCN & $45.2$ & $38.1$ & $82.6$ & $58.1$\tabularnewline

GAT \& ST-GAT & $\mathbf{52.9}$ & $\mathbf{43.2}$ & $\mathbf{88.1}$ & $\mathbf{65.8}$\tabularnewline
\hline 
\end{tabular}
\par\end{centering}

\end{table}
\paragraph{GAT and ST-GAT.} In Table \ref{tab:GAT}, we report the comparison results on VCCR between
the two models using GCN \cite{kipf2017semisupervised} and ST-GCN \cite{yan2018stgcn} and our proposed GAT and ST-GAT for shape and gait learning, respectively. By incorporating GAT and ST-GAT, our proposed model improves Re-ID performance in cloth-changing and standard settings by $5.1\%$ and $7.7\%$ in mAP. Unlike traditional GCN or ST-GCN which operate on graphs by treating every node equally, we leverage attention mechanism, which allows
for attending to local shape cues and local motion ranges, then amplifying
the most important shape features and motion patterns. This helps
mitigates the influence of viewpoint changes and occlusions on shape
and gait, giving more discriminative final person representations.

\begin{table}[ht]
\caption{\label{tab:pose}Ablation study of 3D pose on VCCR.}
\begin{centering}
\small
\begin{tabular}{c|cc|cc}
\hline 
\multirow{2}{*}{Method} & \multicolumn{2}{c|}{CC} & \multicolumn{2}{c}{Standard}\tabularnewline
\cline{2-5} \cline{3-5} \cline{4-5} \cline{5-5} 
 & R-1 & mAP & R-1 & mAP\tabularnewline
\hline 
\hline 
ASGL w/ 2D pose & $47.2$ & $39.9$ & $84.1$ & $60.3$\tabularnewline

ASGL w/ 3D pose  & $\mathbf{52.9}$ & $\mathbf{43.2}$ & $\mathbf{88.1}$ & $\mathbf{65.8}$\tabularnewline
\hline 
\end{tabular}
\par\end{centering}

\end{table}
\paragraph{3D Pose.} In Table \ref{tab:pose}, we provide an insight on the effectiveness
of 3D pose compared to 2D pose in Re-ID. We can observe that using
3D pose leads to an improvement of $3.3\%/5.5\%$ in mAP in cloth-changing/standard
setting. Compared to 2D pose, 3D pose contains richer spatial information. Moreover, 3D pose is less affected by viewpoint changes, thus temporal dynamics from a person's trajectory can be more accurately
captured. The superiority of 3D pose over 2D pose demonstrates the
effectiveness of our ASGL framework for VCCRe-ID task.

\subsection{Discussion}
In Table \ref{ccvid_results}, we also report the performance on CCVID of two models that focus on capturing facial features for Re-ID: (1) the face model InsightFace \cite{Deng2020Retina}; and (2) ReFace \cite{arkushin2022reface}. It can be observed that in standard setting, InsightFace achieves the highest rank-1 accuracy of 95\%, showing that only by extracting facial features, Re-ID performance on CCVID is close to saturation. In cloth-changing setting, ReFace outperforms other works. It is worth noting that ReFace is built upon CAL \cite{Gu2022}, which combines clothes-based loss functions with explicit facial feature extraction. In our future research, we would consider coupling biometric cues like faces with other structural cues like body shape and gait for VCCRe-ID task.

\section{\uppercase{Conclusion}}

In this paper, we proposed Attention-based Shape and Gait Representations Learning, an end-to-end framework for Video-based Cloth-Changing Person Re-Identification. By extracting body shape and gait cues, we enhance the robustness of Re-ID features under clothing-change situations where appearance is unreliable. We proposed a Spatial-Temporal Graph Attention Network (ST-GAT) that encodes gait embedding from 3D-skeleton-based pose sequence. Our ST-GAT is able to amplify important motion ranges as well as capture beneficial local motion patterns for a discriminative gait representation. We showed that by leveraging 3D pose and attention mechanism in our framework, Re-ID accuracy under confusing clothing variations is significantly improved, compared to using 2D pose and traditional Graph Convolutional Networks. Our framework also effectively deals with viewpoint variations and occlusions, shown by state-of-the-art experimental results on the large-scale VCCR dataset which mimics real-world Re-ID scenarios. In future work, we intend to incorporate frame field learning \cite{Nguyen_2024_building} into a multitask learning module to guide the normalization of body pose under viewpoint variations.  
% \section*{\uppercase{Acknowledgements}}

\bibliographystyle{apalike}
{\small
\bibliography{main}}

\end{document}